\documentclass{elsarticle}
\usepackage{graphicx} 
\usepackage{amsmath}
\usepackage{xurl}
\usepackage{hyperref}

\newcommand{\definesuptable}[2]{%
  \expandafter\newcommand\csname #1\endcsname{S#2}%
}

\definesuptable{promptengcomponents}{1}
\definesuptable{pokemonmap}{2}
\definesuptable{kaclabels}{3}
\definesuptable{iaslabels}{4}
\definesuptable{promptengresults}{5}
\definesuptable{macroavg}{6}

\begin{document}

\title{Large language models can disambiguate opioid slang\\ on social media}
\author{Kristy A. Carpenter$^{1}$, Issah A. Samori$^{2}$, Mathew V. Kiang$^{3}$, \\
Keith Humphreys$^{4,8}$, Anna Lembke$^{4}$, Johannes C. Eichstaedt$^{5}$,\\
and Russ B. Altman$^{\dag,1,2,6,7}$}

\address{Departments of $^1$Biomedical Data Science, $^2$Bioengineering, $^3$Epidemiology and Population Health, $^4$Psychiatry and Behavioral Sciences, $^5$Psychology, $^6$Genetics, and $^7$Medicine \\
Stanford University, Stanford, CA 94305, USA \\
$^8$Center for Innovation to Implementation\\
Veterans Affairs Health Care System, Palo Alto, CA 94304, USA\\
$^\dag$E-mail: russ.altman@stanford.edu}

\maketitle

\section{Abstract}
Social media text shows promise for monitoring trends in the opioid overdose crisis; however, the overwhelming majority of social media text is unrelated to opioids. When leveraging social media text to monitor trends in the ongoing opioid overdose crisis, a common strategy for identifying relevant content is to use a lexicon of opioid-related terms as inclusion criteria. However, many slang terms for opioids, such as “smack” or “blues,” have common non-opioid meanings, making them ambiguous. The advanced textual reasoning capability of large language models (LLMs) presents an opportunity to disambiguate these slang terms at scale. We present three tasks on which to evaluate four state-of-the-art LLMs (GPT-4, GPT-5, Gemini 2.5 Pro, and Claude Sonnet 4.5): a lexicon-based setting, in which the LLM must disambiguate a specific term within the context of a given post; a lexicon-free setting, in which the LLM must identify opioid-related posts from context without a lexicon; and an emergent slang setting, in which the LLM must identify opioid-related posts with simulated new slang terms. All four LLMs showed excellent performance across all tasks. In both subtasks of the lexicon-based setting, LLM F1 scores (``fenty'' subtask: 0.824-0.972; ``smack'' subtask: 0.540-0.862) far exceeded those of the best lexicon strategy (0.126 and 0.009, respectively). In the lexicon-free task, LLM F1 scores (0.544-0.769) surpassed those of lexicons (0.080-0.540), and LLMs demonstrated uniformly higher recall. On emergent slang, all LLMs had higher accuracy (average: 0.784), F1 score (average: 0.712), precision (average: 0.981), and recall (average: 0.587) than the two lexicons assessed. The consistency of performance across the four LLMs assessed suggests that these findings generalize across LLMs broadly. Our results show that LLMs can be used to identify relevant content for low-prevalence topics, including but not limited to opioid references, enhancing data provided to downstream analyses and predictive models.

\section{Introduction}
Over 55,000 people in the United States died of an opioid overdose in 2024 \cite{ahmad_provisional_2026} and since the beginning of the opioid crisis, more Americans have died of opioid overdoses than died in World Wars I and II combined \cite{humphreys_responding_2022}. In addition to governmental and academic monitoring of opioid overdose deaths and related metrics of opioid use, social media has emerged as a promising data source from which to obtain real-time insights \cite{sarker_social_2016,smith_monitoring_2025,graves_opioid_2018,vangelov_did_2026}. Content related to opioids, as well as substance use more broadly, appears across various popular, general-purpose social media platforms \cite{carpenter_which_2025,rutherford_turnttrending_2023}. Both explicit opioid mentions \cite{smith_monitoring_2025,chary_epidemiology_2017,sumner_estimating_2022,cuomo_estimating_2023,sarker_machine_2019,graves_opioid_2018,anwar_using_2020} and latent linguistic patterns \cite{giorgi_predicting_2023,matero_opioid_2023} on social media have been shown to correlate with actual opioid overdose death rates.

One key challenge in using social media for monitoring trends in the opioid epidemic is that of “finding a needle in a haystack” – that is, identifying opioid-related content amongst vast quantities of content unrelated to opioids. Manual identification of relevant content done by a domain expert is a common practice for small, qualitative studies \cite{goyer_implementation_2022,hanson_exploration_2013}, but becomes untenable for large-scale quantitative analyses. A common way of filtering for text content of interest is to use a lexicon of opioid terms, excluding any content that does not include any terms from the lexicon \cite{lavertu_redmed_2019,sarker_machine_2019,carpenter_using_2023}.

Assembling a lexicon of opioid terms that may appear on social media is complicated by the casual nature of Internet discussions. Social media content is rife with misspellings, slang, and “algospeak” \cite{moskal_you_2023,steen_you_2023} terms intended to evade automated censorship. Slang terms for opioids are often ambiguous, with common non-drug meanings. This may occur by chance, but this ambiguity is often an intentional way to evade surveillance and censorship. For example, heroin is often referred to as “h” online, but searching for the term “h” on any social media platform will return hundreds of posts completely unrelated to heroin. Lexicon-based methods for identifying opioid-related social media text can either choose to exclude ambiguous terms to prevent the inclusion of widespread noise or to include ambiguous terms to capture additional relevant posts. In the latter case, it is common to manually review posts containing ambiguous terms for opioid relevance before further processing. Manual review is time-consuming, requires access to domain experts who can recognize opioid-related conversation online, and potentially limits the volume of data that can be analyzed.

An alternative to lexicon-based approaches is to extract linguistic features such as Latent Dirichlet Allocation (LDA) topics \cite{blei2003latent}. Although this open-vocabulary approach has proven effective for creating predictive models of mortality \cite{giorgi_predicting_2023,matero_opioid_2023} and thematic exploration \cite{giorgi_cultural_2020}, it requires careful choice of number of topics to model and overall is more difficult to use than lexicon-based and other closed-vocabulary approaches \cite{eichstaedt_closed-_2021}. Furthermore, it may not always be obvious which modeled topics correspond with opioids (or other desired subject matter), making it difficult to use topic modeling approaches to designate posts as opioid-relevant at scale and potentially limiting the type of downstream analysis that can be conducted.

Large language models (LLMs) have experienced widespread breakthroughs in natural language processing tasks, including machine translation \cite{zhu_multilingual_2023}, text summarization \cite{goyal_news_2023,zhang_benchmarking_2023}, and taking exams \cite{bicknell_chatgpt-4_2024}. LLMs, particularly the commonly-used GPT models \cite{brown_language_2020,openai_gpt-4_2024}, have also seen use in analyzing opioid-related social media text. Ge \textit{et al.} found that GPT-3.5 performed well for named entity recognition (NER) of clinical and social impacts of opioid use from posts extracted from opioid-related subreddits \cite{ge_reddit-impacts_2024}. In one study, ChatGPT outperformed statistical, graph-based, and BERT-based \cite{devlin_bert_2019} models in identifying keyphrases from posts on a subreddit dedicated to suboxone \cite{romano_theme-driven_2024}. Hu \textit{et al.} found that ChatGPT outperformed BERT-based baselines in identifying text content related to illicit drug trafficking on Instagram \cite{hu_knowledge-prompted_2024}. Bouzoubaa \textit{et al.} reported that GPT-4 outperformed baseline machine learning models, BERT-based models, and other LLMs in categorizing Reddit posts about personal drug experiences, including those with opioids \cite{bouzoubaa_decoding_2024}. 

We hypothesize that the advanced reasoning capability of LLMs, along with their training on immense corpora of internet text, makes them a promising open-vocabulary approach for classifying social media text.

We introduce a pipeline leveraging various current LLMs to disambiguate potential opioid-related slang terms in social media text. Our pipeline follows from previous work demonstrating the power of GPT models to characterize text previously identified to be likely drug-related (originating from either drug-related subreddits or curated drug-related datasets) and extends it to labeling large tweet datasets for opioid relevance. We demonstrate that all four LLMs evaluated (GPT-4, GPT-5, Claude Sonnet 4.5, and Gemini 2.5 Pro) vastly outperform lexicon-based approaches, suggesting that this ability generalizes across modern LLMs rather than being a feature of any particular model. Our pipeline improves recall of opioid-related social media text without sacrificing precision, allowing researchers to easily increase the amount of relevant social media text upon which to perform a wide range of analyses. While this work focuses on the opioid context, these results ostensibly extend to other low-prevalence and highly stigmatized topics on the internet.

\section{Methods}
\subsection{Model selection}
We evaluated four current commercial LLMs in this study: OpenAI's GPT-4 \cite{openai_gpt-4_2024}, OpenAI's GPT-5 \cite{openai_gpt5_2025}, Anthropic's Claude Sonnet 4.5 \cite{anthropic_sonnet45_2025}, and Google's Gemini 2.5 Pro \cite{comanici_gemini_2025}. We selected these models due to their recency, high performance, similar scale, and accessibility through application programming interface (API).

\subsection{Prompt engineering}
We assembled a set of 100 example tweets from real social media posts to evaluate a variety of prompts and prompting setups. Each of the 100 example tweets was either: a) a tweet from the September 2022 Spritzer dataset (see “Lexicon-based task: Dataset” subsection) containing ambiguous opioid slang terms; b) a tweet about opioids discovered via manual keyword search on Twitter; c) a tweet-length snippet of an r/opiates post on Reddit; d) a tweet containing a drug-related term in a non-opioid-related context, discovered via manual keyword search on Twitter; or e) a totally random non-drug-related tweet (Table 1).

For simplicity, we conducted prompt engineering with GPT-4 only. We iteratively varied different aspects of the prompt scheme presented to GPT-4, optimizing in a greedy manner. We tested 5 different prompts, 2 different temperatures, and 10 different contexts. We tested 6 iterative setups, in which the model is first asked to reason through its answer in free-form before being prompted to respond with a single label of opioid-related, not opioid-related, or unsure for each tweet. Finally, we tested 5 skeptical setups, in which the model is first asked to respond with a single label of  opioid-related, not opioid-related, or unsure for each tweet, then prompted to reconsider its answers and respond again. All prompt components which we used in prompt engineering can be found in Table \promptengcomponents. We evaluated prompt schemes by comparing accuracy, sensitivity, and specificity.

\subsection{Benchmark lexicons}
We selected six lexicons against which to benchmark our LLM-based pipeline: the DEA lexicon \cite{us_drug_enforcement_administration_slang_2018}, the RedMed lexicon \cite{lavertu_redmed_2019}, the Sarker lexicon \cite{sarker_machine_2019}, the Graves lexicon \cite{graves_opioid_2018}, the Yang lexicon \cite{yang_can_2023}, and the Chary lexicon \cite{chary_epidemiology_2017}. We selected these due to their inclusion of ambiguous slang terms for opioids and their previous use for filtering corpora of social media text for content likely to be opioid-related. Although several methods for detecting opioid-related posts on social media have previously been developed \cite{sarker_machine_2019,fodeh_utilizing_2021,al-garadi_large-scale_2022,garg_detecting_2021,chancellor_discovering_2019,preiss_using_2022,yao_detection_2020,al-garadi_text_2021}, none of their code or training data is publicly available.

\subsection{Lexicon-based task}
\subsubsection{Task rationale}
The vast majority of content on general-purpose social media is not related to opioids. With the sheer volume of social media text, using a commercial LLM to query sufficient posts to yield adequate opioid-related signal can become prohibitively expensive and time-consuming. Pre-filtering social media text with a lexicon of opioid-related terms increases prevalence of opioid-related content and decreases total cost, in addition to following precedent from the literature and aiding interpretability. In this task, we assessed whether our GPT-4 workflow could distinguish between instances in which an ambiguous term was used in reference to opioids versus when it was not.
\subsubsection{Dataset: September 2022 Spritzer}
Throughout the 2010s and early 2020s, Twitter made available a Streaming API endpoint that enabled access to new tweets in real-time. Streaming API tiers included the “Firehose” (100\% of tweets), “Gardenhose” (random 10\% of tweets), and “Spritzer” (random 1\% of tweets) variants, which historically have had different price points. The Internet Archive, a non-profit organization aiming to archive digital media across the Internet, makes available historical grabs of the “Spritzer” version of the Twitter stream. We use the September 2022 Spritzer dataset\footnote{https://archive.org/details/archiveteam-twitter-stream-2022-09}, as that is the final month available before substantial changes to the platform occurred as a result of changes in leadership.

The September 2022 Spritzer contains approximately 100 million tweets from September 1 to September 30, 2022 (inclusive). As the Spritzer consists of tweets uniformly selected at random \cite{kergl_endogenesis_2014}, this dataset includes tweets from many different countries and in many different languages. The overwhelming majority of tweets are not opioid-related.

For the lexicon-based task, we chose three ambiguous opioid slang terms: “fenty,” “smack,” and “lean.” Fenty can refer to fentanyl or to a makeup brand; smack can refer to heroin or to the action of hitting something; lean can refer to a drink containing codeine and promethazine \cite{ware_codeine_2024}, to a slant toward something, to the action of resting against something, or to a condition of being thin or slim. We filtered the dataset for any tweet that contained an instance of any of these three terms. There were 492 unique tweets containing “fenty,” 5,895 tweets containing “smack,” and 57,052 tweets containing “lean.”
\subsubsection{Evaluation}
We compared the predicted labels (either “opioid-related,” “not opioid-related,” or “unsure”) from our LLM pipeline to the manually-annotated labels using the accuracy, precision, recall, and F1 score metrics. We calculated the evaluation metrics in two ways: binarizing the task by setting any “unsure” label to “not opioid-related” to be comparable with lexicons; and using multi-class accuracy and macro-averaged precision, recall, and F1 score to capture how well the “unsure” labels aligned between LLM predictions and manual annotations. In either case, any tweet with an error output (either due to prohibited content or tokenizer incompatibility) from the LLM's API was assigned as a “not opioid-related” prediction; this choice was made following the observation that most of the tweets for which GPT-4 produced an error were not opioid-related. We calculated the accuracy, precision, recall, and F1 score when assigning every tweet with the three selected keywords as opioid-related, and when assigning no tweets as opioid-related.

Because there was a prohibitively high number of ``lean'' tweets, we only ran GPT-4 and the lexicon methods on the whole dataset.

\subsection{Lexicon-free task}
\subsubsection{Task rationale}
In cases where running a commercial LLM over an entire social media dataset is feasible, one may not want to be limited by a lexicon, which likely cannot exhaustively list all potential slang terms and misspellings for opioids used on the Internet. We sought to assess if the LLMs could identify opioid-related tweets without being prompted to examine a specific opioid keyword and without conducting preliminary lexicon filtering.
\subsubsection{Dataset: Geolocated tweets}
Geolocating social media posts at the national, state, county, or city level is key for building an informative opioid trend warning system tool. The County Tweet Lexical Bank \cite{schwartz_characterizing_2021} is a dataset with tweets geolocated based on processed location tags and has been previously used for opioid-related work requiring geolocation \cite{giorgi_predicting_2023,matero_opioid_2023}.

We obtained a set of 1628 unique Twitter users geolocated to the states of New York or California from the County Tweet Lexical Bank. We assembled a dataset of approximately 3 million tweets from these users between the dates of 2007 to 2022, which we henceforth refer to as the geolocated dataset. The tweets are predominantly, but not entirely, written in English. The tweets were not filtered for opioid relevance at any point and were expected to be majority unrelated to opioids.

\subsubsection{Evaluation}
We compared the predicted labels (either “opioid-related” or “not opioid-related”) from our LLM pipeline and the six benchmark lexicons to the manually-annotated labels using the accuracy, precision, recall, and F1 score metrics. As in the previous task, we calculated the evaluation metrics in two ways: binarizing the task by setting any “unsure” label to “not opioid-related” to be comparable with the other lexicons; and using multi-class accuracy and macro-averaged precision, recall, and F1 score to capture how well the “unsure” labels aligned between LLM predictions and manual annotations. Again, any tweet with an error output from the LLM's API was assigned as a “not opioid-related” prediction.
\subsection{Emergent slang task}
\subsubsection{Task rationale}
One criticism of lexicon-based identification of opioid-related content is that such methods will become outdated as internet slang evolves, as they will not contain emergent slang terms. This has also been noted as a potential limitation of LLM-based workflows for slang recognition or generation \cite{carpenter_using_2023}, with the reasoning that LLMs will be unable to recognize or generate slang terms not present in their training data. We tested this notion with a third assessment task, in which we provided the LLMs with tweets about opioids that used fake slang terms and assessed whether the LLM could use context to recognize their potential opioid-relevance.
\subsubsection{Dataset: Pokemon}
We selected eight ambiguous slang terms to test for this task: in addition to “fenty,” “smack,” and “lean” from the lexicon-based task, we also used “oxy” (oxycodone, an abbreviation for Occidental College, or an abbreviation for oxygen), “blues” (oxycodone, the color, the genre of music, or the feeling of sadness), “H” (heroin or the letter H), “fetty” (fentanyl or a shortened version of the musician Fetty Wap), and “tar” (heroin or the black substance that is the viscous form of asphalt). We selected 10 tweets each from the September 2022 dataset containing “fenty,” “smack,” and “lean” that were manually determined to be related to opioids. We selected 10 opioid-related tweets for each of “oxy,” “blues,” “H,” “fetty,” and “tar” from the Prompt Engineering dataset and manual keyword search on Twitter. This yielded a total of 80 opioid-related tweets with ambiguous slang terms. We assembled a set of 80 non-opioid-related tweets (10 unique tweets for each of the 8 ambiguous terms) through manual keyword search on Twitter.

We replaced each instance of an ambiguous opioid slang term in both the opioid and non-opioid tweet sets with the name of a Pokemon (a fictional creature from the popular Nintendo game franchise). We used Pokemon names as the fake slang terms because they are often portmanteaus or slight variations of existing words, which is similar to the process by which new slang emerges \cite{grieve_analyzing_2017}. The mapping of ambiguous opioid slang to fake emergent slang terms can be found in Table \pokemonmap.
\subsubsection{Evaluation}
We used our LLM workflow and the six benchmark lexicons to predict whether each tweet in its unmodified form was related to opioids to establish a baseline performance. We repeated this process with the modified tweets. We used the binarized accuracy, precision, recall, and F1 score to evaluate performance.

\subsection{Manual Labeling}
One author (KAC) manually labeled a subset of tweets from both the September 2022 Spritzer and Geolocated CA/NY datasets in order to assess LLM performance (Table \kaclabels). The labeler assigned each tweet an opioid-relevance label: either “opioid-related,” “not opioid-related,” or “unsure.”

All “fenty” and “smack” tweets within the lexicon-based task were manually labeled. The number of “lean” tweets within the lexicon-based task was prohibitively high; the labeler manually labeled all tweets given “opioid-related” (n = 395 tweets) or “unsure” (n = 865 tweets) labels by GPT-4, all tweets for which GPT-4 yielded a Content Restriction Error (n = 987 tweets) or other API Error (n = 52 tweets), and a random 2.25\% (n = 1313 tweets) of all tweets given a “not opioid-related” label by GPT-4. This yielded a total of 3,583 “lean” tweets that were manually labeled, and 57,052 “lean” tweets that were left unlabeled. 

The number of tweets in the Geolocated CA/NY dataset was also prohibitively high for all to be manually labeled. The labeler manually labeled all tweets given “opioid-related” (n = 171 tweets) or “unsure” (n = 391 tweets) labels by GPT-4, a random 1,000 tweets for which GPT-4 yielded a Content Restriction Error, a random 1,000 tweets for which GPT-4 yielded an API Error, and a random 1,500 tweets given a “not opioid-related” label by GPT-4. This yielded a total of 4,062 geolocated tweets that were manually labeled, and 263,583 geolocated tweets that were left unlabeled.

All tweets were stripped of their LLM-generated labels and randomly shuffled before being manually labeled. A subset of 2000 tweets were independently labeled by a second author (IAS) (Table \iaslabels). Interrater agreement was calculated by Cohen's kappa and Spearman correlation. Cohen's kappa measures categorical agreement. We computed Cohen's kappa in two ways: using the ``opioid-related''/``not opioid-related''/``unsure'' labels from each annotator, and binarizing the labels by setting all ``unsure'' labels to ``not opioid-related.'' Spearman correlation measures agreement in ranking, using numerical assignments of 0 for the ``not opioid-related'' label, 1 for the ``unsure'' label, and 2 for the ``opioid-related'' label. The Cohen's kappa, binarized Cohen's kappa, and Spearman correlation between the two manual labelers were 0.5638, 0.6903, and 0.6617, respectively. According to Landis and Koch's widely-used interpretations of the kappa statistic \cite{landis_measurement_1977}, this indicates moderate to substantial agreement. Given the challenging nature of the task and lack of a definite rubric for which tweets are sufficiently opioid-related, this level of agreement is sufficient. All following analyses were therefore conducted against the first annotator's manual labels.

\section{Results}
\subsection{Prompt Engineering}
The best-performing prompt had accuracy of 0.950, sensitivity of 0.870, and specificity of 1.000 on our manually-curated prompt engineering dataset with temperature set to 0 (Table \promptengresults). The context that yielded the best performance was the context used in \cite{carpenter_using_2023}. An iterative prompting scheme in which the model produces a free text response before generating a single label from a constrained vocabulary improved performance to accuracy of 1.000, specificity of 1.000, and sensitivity of 1.000 on the prompt engineering dataset when identifying tweets about opioids in a general sense, and  accuracy of 0.890, specificity of 0.840, and sensitivity of 1.000 when identifying tweets about direct opioid use. Performance decreased when adding a skeptical followup, regardless of the wording of the followup (Table \promptengresults).

For all three tasks and all four LLMs, we use the below iterative prompting setup with temperature set to 0 and otherwise default parameters:
\begin{quote}
Context: “You are an AI assistant that helps people find information. You are particularly hip with online slang and know everything about how people talk on social media platforms like Facebook, Twitter, Reddit, and TikTok.”

User: “I am going to give you a series of tweets, delimited with the xml tags \textless{}tweet\textgreater{}\textless{}/tweet\textgreater{}. For each tweet, I want you to tell me if the tweet is directly referring to opioid use. Reason through your answers step-by-step.

\textless{}tweet\textgreater{} … \textless{}/tweet\textgreater{}

\textless{}tweet\textgreater{} … \textless{}/tweet\textgreater{}

\textless{}tweet\textgreater{} … \textless{}/tweet\textgreater{}”

LLM: …

User: “Based on your reasoning above, answer the question in one word by saying “yes”, “no”, or “unsure” once for each tweet, where “yes” means that the tweet refers to opioids. Separate your answers by commas. Only give this in your response; do not add other content.”

LLM: …
\end{quote}
\subsection{Lexicon-based task}
\begin{figure}[htbp]
    \centering
    \includegraphics[width=\textwidth]{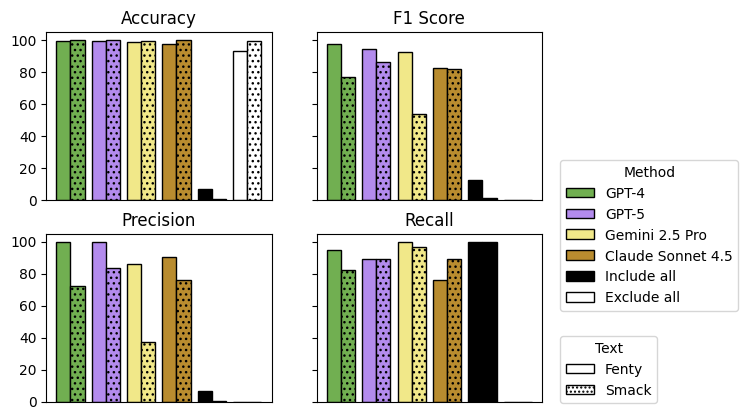}
    \caption{Bar plot of Task 1 evaluation metrics for ``fenty'' (solid bars) and ``smack'' (dotted bars). The four LLMs are compared against the naive benchmarks of including every ``fenty''/``smack'' tweet as opioid-related, or excluding all tweets.}
    \label{fig:disambig_task1_fentysmack_barplot}
\end{figure}

\begin{figure}
    \centering
    \includegraphics[width=\textwidth]{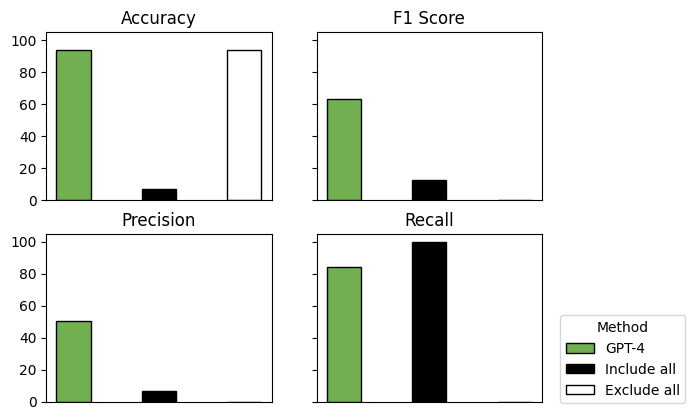}
    \caption{Bar plot of Task 1 (lexicon-based task) evaluation metrics for ``lean'' GPT-4 performance is shown compared against the naive benchmarks of including every ``lean'' tweet as opioid-related, or excluding all tweets.}
    \label{fig:disambig_task1_lean_barplot}
\end{figure}

We determined through manual annotation that of all “fenty” tweets, 6.7\% were opioid-related (33 tweets of 492 tweets total); of all “smack” tweets, 0.5\% were opioid-related (28 tweets of 5895 tweets total); and of all annotated “lean” tweets, 6.6\% were opioid-related (236 tweets of 3583 tweets total). The low prevalence of opioid-related tweets led to low accuracy, precision, and F1 score when assigning all tweets an “opioid-related” label (Figures \ref{fig:disambig_task1_fentysmack_barplot}, \ref{fig:disambig_task1_lean_barplot}), as would happen when including ambiguous terms in a lexicon for social media post inclusion. The low prevalence also led to very high accuracy but 0\% recall when assigning all tweets a “not opioid-related” label (Figures \ref{fig:disambig_task1_fentysmack_barplot}, \ref{fig:disambig_task1_lean_barplot}). In all three cases, the LLM disambiguation workflows had an accuracy similar to or greater than that of the “exclude all” strategy, and their F1 scores (``fenty'': 0.824-0.972; ``smack'': 0.540-0.862; ``lean'': 0.631) were far greater than the F1 score for the “include all” strategy (``fenty'': 0.126; ``smack'': 0.009; ``lean'': 0.124) (the precision and F1 score for the “exclude all” strategy are undefined, as there are no predicted positives) (Figures \ref{fig:disambig_task1_fentysmack_barplot}, \ref{fig:disambig_task1_lean_barplot}). In almost every case, binarizing the LLM (and manual) labels led to better performance metrics than macro-averaging over the “opioid-related,” “unsure,” and “not opioid-related” classes (Table \macroavg).

The four LLMs displayed similar performance on the lexicon-based task, though performance varied more for ``smack'' than for ``fenty'' (Figure \ref{fig:disambig_task1_fentysmack_barplot}). Due to low term prevalence, the accuracies of all four LLMs were all very high and very similar (0.935-0.999 across all subtasks and LLMs). All four LLMs showed a drop in precision and F1 score when applied to ``smack'' tweets as opposed to ``fenty'' tweets, with Gemini 2.5 Pro experiencing the largest performance decline (change in precision: -0.485; change in F1 score: -0.385). Gemini 2.5 Pro performed the best in terms of recall for both ``fenty'' (1.00) and ``smack'' (0.964). The two GPT models performed the best in terms of precision and F1 score for ``fenty'' (GPT-4: precision 1.00, F1 score 0.972; GPT-5: precision 1.000, F1 score 0.943), whereas GPT-5 and Claude Sonnet 4.5 did so for ``smack'' (GPT-5: precision 0.833, F1 score 0.862; Claude: precision 0.758, F1 score 0.820).

All four LLMs had very high precision when labeling the ``fenty'' tweets (Figure \ref{fig:disambig_task1_fentysmack_barplot}), and the GPT models had no false positives (Tables \ref{tab:disambig_gpt4_fenty_cm}, \ref{tab:disambig_gpt5_fenty_cm}). None of the four models produced many ``unsure'' labels for ``fenty'' tweets, with GPT-4 producing 7, Claude Sonnet 4.5 producing 4, Gemini 2.5 Pro producing 1, and GPT-5 producing none (Tables \ref{tab:disambig_gpt4_fenty_cm}-\ref{tab:disambig_claude_fenty_cm}). Only one of these ``unsure'' labelings (by GPT-4) resulted in a false negative. ``Fenty'' tweets manually labeled as ``unsure'' were overwhelmingly labeled by all LLMs as ``not opioid-related.''

The decreased precision (and resulting decreased F1 score) of all four LLMs when labeling the ``smack'' tweets was largely driven by tweets manually labeled as ``opioid-related'' being labeled by the LLM as ``unsure,'' with Gemini 2.5 Pro's larger decrease driven by false negatives from ``not opioid-related'' and ``unsure'' labels equally (Tables \ref{tab:disambig_gpt4_smack_cm}-\ref{tab:disambig_claude_smack_cm}) All four LLMs produced a similar proportion of ``unsure'' labels for ``smack'' as for ``fenty,'' and only two of these ``unsure'' labelings (by the GPT models) resulted in false negatives. ``Smack'' tweets manually labeled as ``unsure'' were overwhelming labeled by all LLMs as either ``not opioid-related'' or ``unsure.'' GPT-4 and Gemini 2.5 Pro both produced errors in response to some of the ``smack'' tweets. Of the ``smack'' tweets for which GPT-4 produced any kind of error, less than 1\% were manually annotated as ``opioid-related.'' All of the ``smack'' tweets for which Gemini 2.5 Pro produced any kind of error were manually annotated as ``not opioid-related.''

Similar patterns followed with GPT-4's performance on the “lean” tweets. 97.2\% of the tweets that GPT-4 classified as “unsure” were manually labeled as either “unsure” or “not opioid-related.” 99.4\% of tweets yielding a Content Restriction Error and 92.3\% of tweets yielding an API Error were manually labeled as “not opioid-related” (Table \ref{tab:disambig_gpt4_lean_cm}).

\subsection{Lexicon-free task}
\begin{figure}
    \centering
    \includegraphics[width=\textwidth]{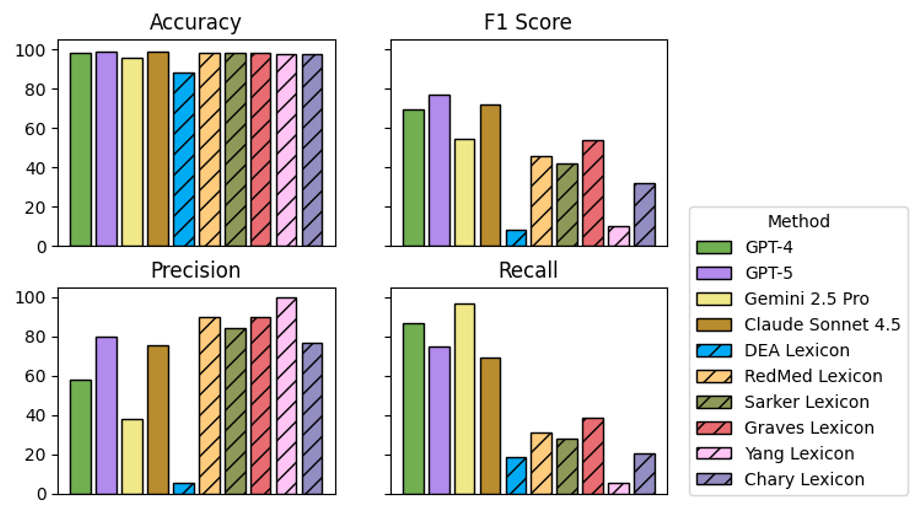}
    \caption{Bar plot of Task 2 (lexicon-free task) evaluation metrics for geolocated data not pre-filtered by a lexicon. The four LLMs (solid bars) are compared against six existing opioid slang lexicons (hatched bars).}
    \label{fig:disambig_task2_barplot}
\end{figure}

Of the 4062 geolocated tweets that were manually labeled, 114 tweets (2.8\%) were manually labeled as ``opioid-related.'' As in the lexicon-based task, low prevalence led to high accuracy for all methods (Figure \ref{fig:disambig_task2_barplot}).

All LLMs had higher recall (0.693-0.965) and F1 score (0.544-0.769) than all lexicons (recall: 0.053-0.386; F1 score: 0.080-0.540) (Figure \ref{fig:disambig_task2_barplot}). The RedMed, Sarker, Graves, and Yang lexicons all had higher precision than all LLMs (0.897, 0.842, 0.898, 1.000, respectively). The Chary lexicon had higher precision (0.767) than all LLMs but GPT-5 (0.794), and the DEA lexicon had by far the worst precision of all methods (0.051).

Among the LLMs, GPT-5 and Claude Sonnet 4.5 had the best F1 score (GPT-5: 0.769; Claude: 0.721) and precision (GPT-5: 0.794; Claude: 0.753), while GPT-4 and Gemini 2.5 Pro had the best recall (GPT-4: 0.868; Gemini: 0.965). Among the lexicons, the Graves lexicon performed the best across metrics, closely followed by the RedMed and Sarker lexicons. The Yang lexicon had the best precision of all methods but otherwise had very poor performance. The DEA lexicon had poor performance overall.

We examined the text of selected tweets to better understand model performance. Some tweets presented below have been slightly modified or shortened.

\begin{equation}
  \tag{Tweet 1}
  \parbox{\dimexpr\linewidth-6em}{%
    \textit{``Illicit fentanyl is incredibly dangerous and has no place on our streets. Whether through enforcement actions or securing funding to combat the opioid epidemic, we are dedicated to protecting our people's health and safety.''}
    }
\end{equation}

Tweet 1 is an example of a tweet that is clearly opioid-related, as it explicitly includes the words ``fentanyl'' and ``opioid''. All LLMs and most lexicons gave Tweet 1 an ``opioid-related'' label. However, the DEA and Yang lexicons did not, as these lexicons do not include the terms ``fentanyl'' or ``opioid.''

\begin{equation}
  \tag{Tweet 2}
  \parbox{\dimexpr\linewidth-6em}{%
    \textit{``And my Norcos kicking in :)''}
    }
\end{equation}
%

In Tweet 2, the term ``norcos'' refers to the brand name of a drug containing hydrocodone, an opioid, and acetaminophen. This term was only included in the RedMed lexicon, leading to the other five lexicons labeling Tweet 2 as ``not opioid-related.'' All four LLMs labeled this tweet as opioid-related.

\begin{equation}
  \tag{Tweet 3}
  \parbox{\dimexpr\linewidth-6em}{%
    \textit{``2 Chainz sound weak AF on that drank in my cup remix''}
    }
\end{equation}

Tweet 3 is an example of a tweet which received different labels from different LLMs: GPT-4 and Gemini gave it an ``opioid-related'' label, while GPT-5 and Claude gave it a ``not opioid-related'' label. In this tweet, ``drank in my cup'' refers to a song titled for lean, a drink containing the opioid codeine. While a tweet about lean would be considered opioid-related, it is unclear whether a tweet about a song about lean is opioid-related, leading to the discrepancies in LLM labeling. All lexicons gave Tweet 3 a ``not opioid-related'' label.

\begin{equation}
  \tag{Tweet 4}
  \parbox{\dimexpr\linewidth-6em}{%
    \textit{``2 good 2 be true!! rip actavis''}
    }
\end{equation}

Finally, Tweet 4 is an example of a tweet for which the LLM labels unanimously disagreed with the manual label. The manual labeler and all lexicons assigned this tweet a label of ``not opioid-related.'' However, all four LLMs assigned this tweet a label of ``opioid-related.'' Upon closer inspection, we determined that ``actavis'' refers to a pharmaceutical company which formerly produced promethazine with codeine cough syrup, the primary ingredient of lean. With this context, we determined that Tweet 4 likely is opioid-related, despite the initial manual annotation otherwise.

\subsection{Emergent slang task}
\begin{figure}
    \centering
    \includegraphics[width=\textwidth]{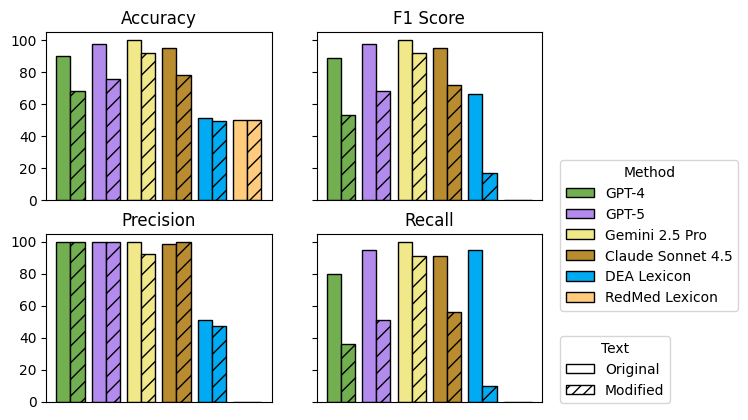}
    \caption{Bar plot of Task 3 (emergent slang task) evaluation metrics. Performance is shown for both the unmodified tweets (solid bars) and modified tweets (hatched bars). The four LLMs are compared against two existing opioid slang lexicons.}
    \label{fig:disambig_task3_barplot}
\end{figure}

All four LLMs performed very well on the dataset of unmodified tweets, with accuracies ranging from 0.900 to 1.000 and F1 scores ranging from 0.889 to 1.000 (Figure \ref{fig:disambig_task3_barplot}). Reflecting the balanced nature of the dataset, both lexicons assessed had an accuracy of about 0.500; the DEA lexicon assigned all tweets in the dataset as ``opioid-related,'' leading to precision of 0.500 and recall of 1.000, whereas the RedMed lexicon assigned all tweets as ``not opioid-related,'' leading to undefined precision and recall of 0.000. We omitted the remaining four lexicons as any lexicon-based approach would fall into one of these two result patterns by the design of the dataset.

After modifying the tweets to simulate emergent slang, accuracy, F1 score, and recall dropped for all LLMs (Figure \ref{fig:disambig_task3_barplot}). However, precision remained high, ranging from 0.924 to 1.000. While recall markedly decreased from the original data, all LLMs maintained a recall greater than 0.500. Modification of the dataset led to the DEA lexicon's recall dropping from 0.950 to 0.100, in turn causing the F1 score to drop dramatically. The tweets classified as opioid-related after modification contained additional terms included in the DEA lexicon besides those targeted for modification. These additional terms did not provide adequate signal to determine opioid relevance, as evidenced by the precision of the DEA lexicon on the modified tweets staying around 0.500. As the RedMed lexicon did not assign any tweets the ``opioid-related'' label before modification, modifying the tweets did not alter performance. 

\section{Discussion}
This study examined whether LLMs could filter large volumes of social media text data to identify posts related to opioids. In a lexicon-based setting, a lexicon-free setting, and an emergent slang setting, all four LLMs assessed were successful in accurately distinguishing between opioid-related and non-opioid-related content. While the performance of individual LLMs varied over different tasks, they all consistently outperformed the lexicon-based approaches, demonstrating the utility of commercial-scale LLMs generally rather than that of one particular LLM. This study focused on identifying opioid-related text on social media, but the presented LLM framework could easily be applied to identifying social media content related to other low-prevalence topics.

The key contribution of LLMs in the task of identifying relevant content for social media-based monitoring of the opioid crisis is their increased recall. While lexicons require brand names of opioids, generic names of opioids, slang terms for opioids, misspellings, and algospeak to be exhaustively prespecified in order to capture low-prevalence opioid content, the open-vocabulary approach and contextual reasoning ability of LLMs allow them more flexibility. This was especially evident in Task 2 (lexicon-free classification), where the LLM models had markedly higher recall than the lexicons and the example tweets included instances of seemingly obvious terms being left out of lexicons. Increased recall at times comes at the cost of reduced precision. However, in this setting, high recall is preferable to high precision; using the LLM as an initial filtering step makes a subsequent manual review phase easier and higher yield.

Increased recall is particularly relevant to the emergent slang case. Even if the LLMs do not identify all tweets with novel slang, identifying a few examples could allow researchers to recognize a new slang term to include in lexicons or search for in a large corpus.

Some tweets were caught by the LLMs' content restriction filters or otherwise produced errors. We expected that content related to illicit drug use might trigger a content restriction response, but we found that tweets flagged for content violation were overwhelmingly not related to opioids. Instead, these tweets were flagged for being graphically violent, graphically sexual, or otherwise objectionable. Tweets that yielded other errors largely consisted of non-unicode characters, mojibake \cite{erickson_plain_2021}, or external links, indicating that the errors were related to processing the input text and that most, if not all, tweets with this error would be unrelated to opioids.

Our analysis assumed that the manual annotations from a human labeler are ground truth. However, lack of knowledge of a particular slang word or piece of media as well as simple human error are almost guaranteed to have introduced mistakes in the human labels. We included an example of such an error with Tweet 4, which was incorrectly manually assigned a ``not opioid-related'' label. This suggests that beyond being faster than manual review, LLMs may be more accurate than manual review. This also suggests that there may be other unnoticed instances of human error and that LLM performance may be slightly better than the results presented here.

This study has some limitations. We only conducted prompt engineering with GPT-4 and used the same context and prompt setup across all four LLMs evaluated. Performance could possibly have improved for GPT-5, Claude Sonnet 4.5, or Gemini 2.5 Pro with additional prompt engineering using those models. Given that all models showed excellent performance across tasks and that prompt engineering can be conducted endlessly, we believe that further prompt engineering for this study was not warranted. However, in a deployment context, it may be necessary to revisit individual model performance to make choices such as if an ``unclear'' label gets pushed to ``opioid-related'' or ``not opioid-related.''

Another limitation of this study is that it was not feasible to manually annotate all tweets presented to the LLMs. While we selected our subsets for manual labeling in a manner aiming to yield the most insight to model performance, evaluating against ground truth labels for the entire dataset may slightly change the resulting metrics.

A final limitation of this study is that while the presented workflow is easy to use, it is not without cost. Those wishing to leverage commercial LLMs to identify opioid-related social media text, as we have demonstrated, will need to pay the companies providing the LLMs and may find the cost prohibitively high. Free LLMs are available for use on HuggingFace, such as Llama 3 and DeepSeek V3, but models of comparable size to the models tested here require access to significant GPU compute.

We recognize that applying LLMs to identify opioid-related text on social media platforms could lead to surveillance at the individual level. We only condone monitoring at the population level to inform and target harm reduction, prevention, and treatment strategies. We do not condone this application and strongly oppose using LLMs for automated censorship, profiling, or criminalization of individuals deemed to be potentially using opioids. Besides being unethical, it is not possible to accurately detect whether an individual is using opioids or not from their social media posts with an LLM or any other text analysis method, as any statement posted on the internet may be entirely false.

\section{Supplementary Information}
\subsection{Availability of data and materials}
Code is available at \url{https://github.com/Helix-Research-Lab/opioid-disambiguation}. Given the potentially sensitive nature of the data at hand, combined with the users' lack of informed consent about inclusion in an opioid-associated dataset, the social media text data will only be provided upon reasonable request.

\subsection{Acknowledgments}
This work supported by: NIH DA057598; Microsoft Accelerating Foundation Models Research Initiative; NSF GRFP DGE-1656518 to KAC; Sarafan ChEM-H CBI Program Award to IAS; KH is supported by a Senior Research Career Scientist Award (RCS 04-141-3) from the Department of Veterans Affairs Health System Research Service and a grant from the National Institute on Drug Abuse (2UG1DA015815-19); JCE, RBA, AL are supported by the Stanford Institute for Human-Centered AI; NIH GM153195 to RBA; RBA is supported by the Chan Zuckerberg Biohub. We thank Betty Xiong, Delaney Smith, Anna Nguyen, and Aadesh Salecha for informative discussion. We thank Shashanka Subrahmanya for assistance with the geolocated dataset.

\subsection{Competing Interests}
The authors have no competing interests to declare.

\subsection{Author Contributions}
Conceptualization: K.A.C., J.C.E., R.B.A.; Methodology: K.A.C., J.C.E., R.B.A.; Software: K.A.C.; Validation: K.A.C.; Investigation: K.A.C.; Resources: K.A.C., J.C.E.; Data Curation: K.A.C., I.A.S., J.C.E.; Writing - Original Draft: K.A.C.; Writing - Review \& Editing: K.A.C., I.A.S., M.V.K., A.L., K.H., J.C.E, R.B.A.; Visualization: K.A.C.; Supervision: M.V.K., K.H., A.L., J.C.E., R.B.A.; Project administration: R.B.A.; Funding acquisition: I.A.S., M.V.K., K.H., A.L., J.C.E., R.B.A

\newpage
\section{Figures}

\newpage
\section{Tables}

\begin{table}[htbp]
    \centering
    \begin{tabular}{lcccc}
          &&  \multicolumn{3}{c}{Manual label}\\
          &&  Opioid-related&  Not opioid-related&  Unsure\\
          &Opioid-related&  35&  0&  0\\
          GPT-4 label&Not opioid-related&  1&  412&  37\\
  &Unsure& 1& 3& 3\\
    \end{tabular}
    \caption{Confusion matrix for GPT-4 labeling of ``fenty'' tweets.}
    \label{tab:disambig_gpt4_fenty_cm}
\end{table}

\begin{table}
    \centering
    \begin{tabular}{lcccc}
          &&  \multicolumn{3}{c}{Manual label}\\
          &&  Opioid-related&  Not opioid-related&  Unsure\\
          &Opioid-related&  33&  0&  0\\
          GPT-5 label&Not opioid-related&  4&  415&  40\\
  &Unsure& 0& 0& 0\\
    \end{tabular}
    \caption{Confusion matrix for GPT-5 labeling of ``fenty'' tweets.}
    \label{tab:disambig_gpt5_fenty_cm}
\end{table}

\begin{table}
    \centering
    \begin{tabular}{lcccc}
          &&  \multicolumn{3}{c}{Manual label}\\
          &&  Opioid-related&  Not opioid-related&  Unsure\\
          &Opioid-related&  37&  2&  4\\
          Gemini label&Not opioid-related&  0&  413&  35\\
  &Unsure& 0& 0& 1\\
    \end{tabular}
    \caption{Confusion matrix for Gemini 2.5 Pro labeling of ``fenty'' tweets.}
    \label{tab:disambig_gemini_fenty_cm}
\end{table}
\begin{table}
    \centering
    \begin{tabular}{lcccc}
          &&  \multicolumn{3}{c}{Manual label}\\
          &&  Opioid-related&  Not opioid-related&  Unsure\\
          &Opioid-related&  28&  2&  1\\
          Claude label&Not opioid-related&  9&  412&  36\\
  &Unsure& 0& 1& 3\\
    \end{tabular}
    \caption{Confusion matrix for Claude Sonnet 4.5 labeling of ``fenty'' tweets.}
    \label{tab:disambig_claude_fenty_cm}
\end{table}
\begin{table}
    \centering
    \begin{tabular}{lcccc}
          &&  \multicolumn{3}{c}{Manual label}\\
          &&  Opioid-related&  Not opioid-related&  Unsure\\
          &Opioid-related&  23&  0&  9\\
          &Not opioid-related&  2&  5370&  44\\
  GPT-4 label&Unsure& 1& 86& 42\\
 & Content Restriction Error& 0& 262&5\\
 & API Error& 2& 47&2\\
    \end{tabular}
    \caption{Confusion matrix for GPT-4 labeling of ``smack'' tweets.}
    \label{tab:disambig_gpt4_smack_cm}
\end{table}

\begin{table}
    \centering
    \begin{tabular}{lcccc}
          &&  \multicolumn{3}{c}{Manual label}\\
          &&  Opioid-related&  Not opioid-related&  Unsure\\
          &Opioid-related&  25&  0&  5\\
          GPT-5 label&Not opioid-related&  2&  5709&  61\\
  &Unsure& 1& 56& 36\\
    \end{tabular}
    \caption{Confusion matrix for GPT-5 labeling of ``smack'' tweets.}
    \label{tab:disambig_gpt5_smack_cm}
\end{table}

\begin{table}
    \centering
    \begin{tabular}{lcccc}
          &&  \multicolumn{3}{c}{Manual label}\\
          &&  Opioid-related&  Not opioid-related&  Unsure\\
          &Opioid-related&  27&  20&  25\\
          Gemini label&Not opioid-related&  1&  5737&  73\\
  &Unsure& 0& 2& 4\\
 & API Error& 0& 6&0\\
    \end{tabular}
    \caption{Confusion matrix for Gemini 2.5 Pro labeling of ``smack'' tweets.}
    \label{tab:disambig_gemini_smack_cm}
\end{table}
\begin{table}
    \centering
    \begin{tabular}{lcccc}
          &&  \multicolumn{3}{c}{Manual label}\\
          &&  Opioid-related&  Not opioid-related&  Unsure\\
          &Opioid-related&  25&  1&  7\\
          Claude label&Not opioid-related&  3&  5742&  73\\
  &Unsure& 0& 22& 22\\
    \end{tabular}
    \caption{Confusion matrix for Claude Sonnet 4.5 labeling of ``smack'' tweets.}
    \label{tab:disambig_claude_smack_cm}
\end{table}

\begin{table}
    \centering
    \begin{tabular}{lccccl}
          &&  \multicolumn{3}{c}{Manual label} &\\
          &&  Opioid-related&  Not opioid-related&  Unsure &Unlabeled\\
          &Opioid-related&  199&  36&   160&0\\
          &Not opioid-related&  7&  1229&   48&57052\\
  GPT-4 label&Unsure& 24& 537&  304&0\\
 & Content Restriction Error& 3& 981& 3&0\\
 & API Error& 3& 48& 1&0\\
    \end{tabular}
    \caption{Confusion matrix for GPT-4 labeling of ``lean'' tweets.}
    \label{tab:disambig_gpt4_lean_cm}
\end{table}
\newpage
\bibliographystyle{unsrt}
\bibliography{mybib}

\end{document}